\documentclass[11pt]{article} %

\usepackage{rldm,palatino}
\usepackage{graphicx}
\usepackage[colorlinks=true,linkcolor=blue,citecolor=blue,urlcolor=blue]{hyperref}
\usepackage{amscd,amsfonts,amssymb,amstext,latexsym}  %
\usepackage{amsmath,mathbbol,mathrsfs,stmaryrd, mathtools}  %
\usepackage[numbers,sort&compress]{natbib}
\usepackage{caption}
\usepackage{subcaption}
\usepackage{algorithm}
\usepackage{algpseudocode}
\usepackage{todonotes}
\usepackage[colorlinks=true,linkcolor=blue,citecolor=blue,urlcolor=blue]{hyperref}
\usepackage{cleveref}

\usepackage{graphicx}

\usepackage{amscd,amsfonts,amssymb,amstext,latexsym} %
\usepackage{amsmath,mathbbol,mathrsfs,stmaryrd, mathtools} %
\usepackage{amsmath}
\usepackage{amssymb}
\usepackage{amsthm}

\usepackage[numbers,sort&compress]{natbib}
\usepackage{caption}
\usepackage{algorithm}
\usepackage{algpseudocode}
\usepackage{todonotes}  %
\usepackage{booktabs}  %
\usepackage{makecell}  %

\usepackage{graphicx,wrapfig,adjustbox,lipsum}
\usepackage[inline]{enumitem}
\usepackage{amsthm}

\usepackage{mathtools}

\DeclarePairedDelimiterX{\expectarg}[1]{[}{]}{%
  \ifnum\currentgrouptype=16 \else\begingroup\fi
  \activatebar#1
  \ifnum\currentgrouptype=16 \else\endgroup\fi
}

\DeclarePairedDelimiterX{\probarg}[1]{(}{)}{%
  \ifnum\currentgrouptype=16 \else\begingroup\fi
  \activatebar#1
  \ifnum\currentgrouptype=16 \else\endgroup\fi
}

\newcommand{\innermid}{\nonscript\;\delimsize\vert\nonscript\;}
\newcommand{\activatebar}{%
  \begingroup\lccode`\~=`\|
  \lowercase{\endgroup\let~}\innermid 
  \mathcode`|=\string"8000
}

\title{Accounting for the Sequential Nature of States to Learn \\ Features for Reinforcement Learning}
\author{
Nathan Michlo, Devon Jarvis, Richard Klein, Steven James \\
School of Computer Science and Applied Mathematics\\
University of the Witwatersrand\\
Johannesburg, South Africa \\
\texttt{\{nathan.michlo1, devon.jarvis1\}@students.wits.ac.za}\\
\texttt{\{richard.klein,steven.james\}@wits.ac.za} \\
}

\newlength{\strutheight}
\usepackage{amsmath,amsfonts,bm}

\def\eqref#1{equation~\ref{#1}}

\def\1{\bm{1}}

\DeclareMathAlphabet{\mathsfit}{\encodingdefault}{\sfdefault}{m}{sl}
\SetMathAlphabet{\mathsfit}{bold}{\encodingdefault}{\sfdefault}{bx}{n}

\usepackage{calc}
\usepackage{amsmath}
\usepackage{mathtools}
\usepackage{mathrsfs}
\usepackage{array}

\newcommand{\bb}[1]{\boldsymbol{#1}}  %
\newcommand{\set}[1]{\mathcal{#1}}

\renewcommand{\i}[1][]{\def\temp{#1}\ifx\temp\empty {} \else ^{(#1)} \fi}

\newcommand{\bx}[1][]{\bb{x}\i[#1]}
\newcommand{\br}[1][]{\hat{\bb{x}}\i[#1]}
\newcommand{\bz}[1][]{\bb{z}\i[#1]}
\newcommand{\by}[1][]{\bb{y}\i[#1]}

\newcommand{\btheta}{\boldsymbol{\theta}}
\newcommand{\bphi}{\boldsymbol{\phi}}
\newcommand{\beps}{\boldsymbol{\epsilon}}

\newcommand{\bmu}[1][]{\boldsymbol{\mu}\i[#1]}
\newcommand{\bsigma}[1][]{\boldsymbol{\sigma}\i[#1]}

\newcommand{\X}{\mathbf{\mathcal{X}}}  %
\newcommand{\sX}{\mathrm{N}}  %
\newcommand{\sZ}{\mathrm{D}}  %

\newcommand{\R}{\mathbf{\mathbb{R}}}

\newcommand{\Enc}{f_{\bphi}}
\newcommand{\Dec}{g_{\btheta}}

\newcommand{\qEnc}{q_{\bphi}}       %
\newcommand{\pDec}{p_{\btheta}}     %
\newcommand{\pTrue}{p_{\ast}}       %

\NewDocumentCommand{\argZX}{O{}O{}}{\def\temp{#1}\ifx\temp\empty {(\bz|\bx[#2])} \else {(z_#1|\bx[#2])} \fi}
\NewDocumentCommand{\argXZ}{O{}O{}}{\def\temp{#1}\ifx\temp\empty {(\bx|\bz[#2])} \else {(x_#1|\bz[#2])} \fi}

\NewDocumentCommand{\qzx}{O{}O{}}{\qEnc\argZX[#1][#2]}  %
\NewDocumentCommand{\pxz}{O{}O{}}{\pDec\argXZ[#1][#2]}  %
\NewDocumentCommand{\pz}{O{}}{\pDec(\bz)}        %
\newcommand{\txz}{\pTrue(\bx|\bz)}  %
\newcommand{\tz}{\pTrue(\bz)}       %

\newcommand{\Loss}{\mathcal{L}}

\newcommand{\LossReg}{\Loss_{\mathrm{reg}}}
\newcommand{\LossRec}{\Loss_{\mathrm{rec}}}
\newcommand{\LossVAE}{\Loss_{\mathrm{VAE}}}
\newcommand{\LossTVAE}{\Loss_{\beta\mathrm{TVAE}}}

\newcommand{\LossBVAE}{\Loss_{\beta\mathrm{VAE}}}

\newcommand{\LossAdaTrip}{\Loss_{\mathrm{Ada-Triplet}}}

\newcommand{\braces   }[1]{\left\{#1\right\}}
\newcommand{\brackets }[1]{\left[#1\right]}
\newcommand{\parens   }[1]{\left(#1\right)}

\newcommand{\norm}[1]{\lVert #1 \rVert}

\newcommand{\E}{\mathbb{E}}

\newcommand{\Dkl}{D_{\mathrm{KL}}}

\newcommand{\Dist}{\mathrm{d}}

\newcommand{\VDistance}{\Dist_\mathrm{vis}}
\newcommand{\GtDistance}{\Dist_\mathrm{gt}}

\newcommand{\elemprod}{\odot}

\newcolumntype{P}[1]{>{\centering\arraybackslash}p{#1}}

\makeatletter
\providecommand*{\barvee}{%
  \mathbin{%
    \mathpalette\@barvee{}%
  }%
}
\newcommand*{\@barvee}[2]{%
  \sbox0{$#1\veebar\m@th$}%
  \sbox2{%
    \hbox to \wd0{%
      \hss
      \resizebox{1.05\wd0}{\height}{$#1-\m@th$}%
      \hss
    }%
  }%
  \sbox4{%
    \resizebox{\wd0}{.7\ht0}{$#1\vee\m@th$}%
  }%
  \sbox6{$#1\vcenter{}$}
  \ht2=\ht6 %
  \vbox to \ht0{%
    \copy2 %
    \vss
    \copy4 %
  }%
}
\makeatother

\begin{document}

\maketitle

\begin{abstract}

In this work, we investigate the properties of data that cause popular representation learning approaches to fail. In particular, we find that in environments where states do not significantly overlap, variational autoencoders (VAEs) fail to learn useful features. We demonstrate this failure in a simple gridworld domain, and then provide a solution in the form of metric learning. However, metric learning requires supervision in the form of a distance function, which is absent in reinforcement learning. To overcome this, we leverage the sequential nature of states in a replay buffer to approximate a distance metric and provide a weak supervision signal, under the assumption that temporally close states are also semantically similar. We modify a VAE with triplet loss and demonstrate that this approach is able to learn useful features for downstream tasks, without additional supervision, in environments where standard VAEs fail.

\end{abstract}

\keywords{
representation learning, unsupervised learning, metric learning, reinforcement learning
}

\startmain %

\section{Introduction}\label{sec:intro}
    
    A fundamental challenge in machine learning is to discover useful representations from high-dimensional data, which can then be used to solve subsequent tasks effectively. 
    Recently, deep learning approaches have showcased the ability of neural networks to extract meaningful features from high-dimensional inputs for tasks in both the supervised and reinforcement learning (RL) setting \citep{krizhevsky2012imagenet,mnih2015human}. 
    While these learnt representations have obviated the need for manual feature selection or preprocessing pipelines, they are often not semantically meaningful, which can negatively impact downstream task performance~\citep{locatello2018challenging}.  
    Prior work has therefore argued that it is desirable to learn a representation that is \textit{disentangled}. %
    While there is no consensus on what constitutes a disentangled representation, it is generally agreed that such a representation should be factorised so that each latent variable corresponds to a single explanatory variable responsible for generating the data \citep{burgess2018understanding}.
    For example, a single image from a video game may be represented by latent variables governing the $x$ and $y$ positions of the player, enemies and collectable items.

    In deep RL, there are two general approaches to feature learning. The first is to implicitly discover features by simply training an agent to solve a given task, relying on a neural network function approximator and stochastic gradient descent to uncover useful features.
    An alternative approach is to explicitly learn features through some auxiliary objective. 
    One such method is to use variational autoencoders (VAEs) \cite{kingma2013auto}, which are trained on collected data to learn a lower-dimensional representation capable of reconstructing the given input. 
    VAEs have been shown to produce disentangled representations when trained on synthetically generated data, although there is still no explicit reason why these representations should align with generative factors in the data.

    In this work, we investigate the correspondence between the VAE reconstruction loss and learnt distances in the latent space. We demonstrate the ineffectiveness of state-of-the-art models by designing a simple visual gridworld domain over which VAE-based frameworks fail to learn useful features for a downstream RL task. We overcome this limitation by utilising the sequential nature of an RL replay buffer along with metric learning to guide representations learnt by VAEs. Furthermore, we design a disentangled metric learning approach which further improves learnt representations for downstream tasks.

\section{Background}
    
    We model an agent acting in an environment by a Markov decision processes (MDP) $\mathcal{M} = \langle S, A, T, R, \gamma \rangle$ where 
    \begin{enumerate*}[label=(\roman*)]
    \item $S$ is the state space;
    \item $A$ is the set of actions;
    \item $T(s, a, s')$ describes the transition dynamics, specifying the probability of arriving in state $s'$ after action $a$ is executed from $s$; 
    \item $R(s, a)$ specifies the reward for executing action $a$ in state $s$; and
    \item $\gamma \in [0, 1)$ is a discount factor.
    \end{enumerate*}
    The agent's aim is to compute a policy $\pi$ that maps from states to actions and optimally solves the task.
    Given a policy $\pi$, RL algorithms often estimate a value function $v^\pi(s) = \mathbb{E}_\pi \big[ \sum_{i=0}^{\infty} \gamma^i R(s_{i}, a_{i}) | s_0 = s \big]$, representing the expected return obtained following $\pi$ from state $s$.
    The \textit{optimal} policy $\pi^*$ is the policy that obtains the greatest expected return at each state: $v^{\pi^*}(s) = v^*(s) = \max_\pi v^\pi(s)$ for all $s \in S$.
    A related quantity is the action-value function, $q^\pi(s, a)$, which defines the expected return obtained by executing $a$ from $s$, and thereafter following $\pi$.
    Similarly, the optimal action-value function is given by $q^*(s, a) = \max_\pi q^\pi(s, a)$ for all states $s$ and actions $a$ \citep{sutton2018reinforcement}.
    
    For high-dimensional or continuous state spaces, representing the value function precisely is intractable. 
    Instead, the value function is normally approximated through a parameterised function $v^\pi(s) \approx \hat{v}(s; \mathbf{w})$, where $\mathbf{w}$ is some parameter vector to be learned; in deep RL, these parameters are the weights of a neural network that can be trained to approximate the value function.
    Since these neural networks require significant amounts of data to train, one way to improve sample efficiency is to use a \textit{replay buffer}, where old experience data is stored in memory and reused to update the network as it continues to interact with the environment. 

\subsection{Representation Learning}

        Assume a dataset $\X = \braces{\bx[0], ..., \bx[n]}$ is a set of independent and identically distributed (i.i.d)%
        \footnote{In RL, a replay buffer is not i.i.d, as data is generated by a sequential decision process. This can be mitigated via random sampling.}
        observations $\bx \in \R^\sX$, generated by some random process involving an unobserved random variable $\bz \in \R^\sZ$ of lower dimensionality $\sZ \ll \sX$.
        Additionally, the true \textit{prior distribution} $\bz \sim \tz$ and true \textit{conditional distribution} $\bx \sim \txz$ are unknown. 
        Variational autoencoders (VAEs) aim to learn this generative process. 
        Unlike autoencoders (AEs), which consist of an encoder $\Enc(\bx) = \bz$ and decoder $\Dec(\bz) = \br$ with weights $\bphi$ and $\btheta$, VAEs instead construct a probabilistic encoder by using the output from the encoder or inference model to parameterise approximate posterior distributions $\bz~\sim~\qzx$. 
        The approximate posterior is then sampled from during training to obtain representations $\bz$, which are then decoded using the generative model to obtain reconstructions $\br~\sim~\pxz$.
        
        A \textit{factorised Gaussian encoder} \citep{kingma2013auto} is commonly used to model the posterior using a multivariate Gaussian distribution with diagonal covariance $\bz \sim \set{N}(\bmu_{\bphi}(\bx),\;\bsigma_{\bphi}(\bx))$, with the prior given by the multivariate normal distribution $\pz = \set{N}(\mathbf{0},\;\mathbf{I})$, with a mean of $\mathbf{0}$ and diagonal covariance $\mathbf{I}$. 
        To enable backpropagation, the reparameterisation trick in \Cref{eq:reparameterization-trick} is used to sample from the posterior distribution during training by offsetting the distribution means by scaled noise values.
        \begin{align}
            \beps \sim \set{N}(\mathbf{0}, \mathbf{I}); \quad  \bz = \bmu_{\bphi}(\bx) + \bsigma_{\bphi}(\bx) \odot \beps \label{eq:reparameterization-trick}
        \end{align}

        VAEs maximise the evidence lower bound (ELBO) by minimising the loss given by \Cref{eq:vae}. 
        VAE-based approaches often make slight modifications to this loss but generally the terms can still be grouped into regularisation and reconstruction components.
        The regularisation term constrains the representations learnt by the encoder, while the reconstruction term improves the outputs of the decoder so that they better match the inputs to the encoder. 
        These terms usually conflict in practice, strong regularisation leads to worse reconstructions but often better disentanglement \cite{higgins2017beta,burgess2018understanding}. %
        \begin{align}
            \LossRec(\bx, \br) = \E_{\qzx} \brackets{\log \pxz}; \quad
            \LossReg(\bx) = - \Dkl \parens{\qzx \;\|\; \pz}; \quad
            \LossVAE(\bx, \br) = \LossRec(\bx, \br) + \LossReg(\bx) \label{eq:vae}
        \end{align}

\section{A Motivating Gridworld Environment} \label{section:adverserial-dataset}

\begin{figure*}[h!]
        \centering
            \begin{subfigure}{0.3\linewidth}
                \includegraphics[width=1.0\linewidth]{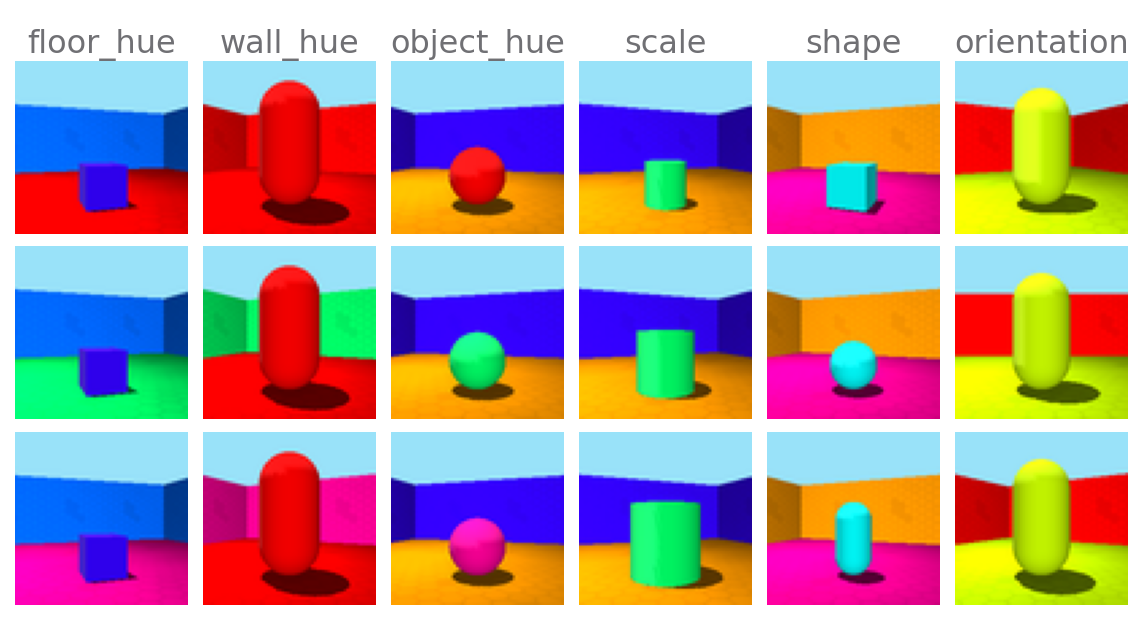}
                \caption{3D~Shapes~\cite{3dshapes18}}\label{subfig:3dshapes}
            \end{subfigure}
        \quad
            \begin{subfigure}{0.5\linewidth}
                \includegraphics[width=1.0\linewidth]{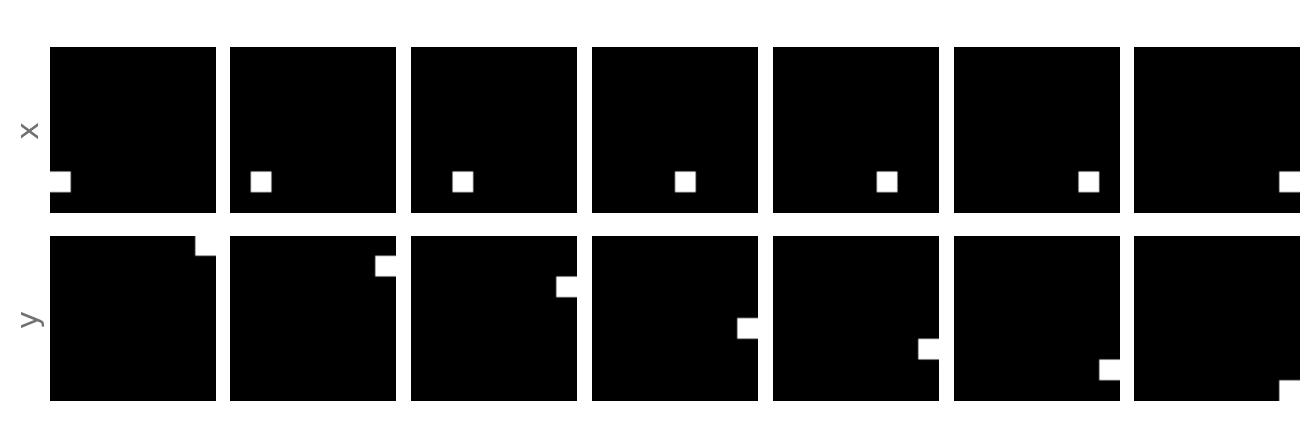}
                \caption{Simple gridworld domain with pixel-based states}
                \label{subfig:gridworld} \label{fig:data-traversal_xysquares}
            \end{subfigure}

        \caption{
            (a) A common synthetic dataset---images are generated by smoothly varying ground-truth factors such as floor hue and object shape. (b) Gridworld environment whose underlying factors are the agent's $xy$-position.
        }
        \label{fig:data-traversals_shapes3d-cars3d}
        \label{fig:data-traversals_all}
    \end{figure*}

    Before investigating representation learning in RL, we first consider the simpler unsupervised learning setting, in which VAEs and their variants are primarily developed through benchmarks over synthetic data.  
    The 3D Shapes dataset \cite{3dshapes18} in \Cref{subfig:3dshapes} contains observations of shapes fixed in the centre of the image with progressively changing attributes or factors such as size and colour. If, as humans, we are given unordered observations from a traversal along the size factor of 3D~Shapes, it would be easy to order these observations using a perceived increase or decrease in the size of the shape. We might even say that the shapes in the images overlap by different amounts. 
    We may consider shapes that are closer in size to possess more overlap, and therefore also consider them to be closer together in terms of distance.
    In contrast, consider a simple gridworld environment (Figure~\ref{subfig:gridworld}) where an agent moves to adjacent cells in any of the four cardinal directions. 
    The agent is a square of size $8 \times 8$ pixels and each action translates it 8 pixels in a given direction. 
    Note that in this case, there is no overlap between the agent in any two images because of the distance it moves.

    It is therefore natural to ask whether a lack of overlap is problematic. 
    To answer this question, let $\bx$ be an observation and $\by$ be the underlying factors that generated it. 
    We define the \textit{ground-truth distance} between pairs of observations $\{\bx[a], \bx[b]\}$ as the Manhattan distance between the observations' underlying factors:  $\GtDistance(\bx[a], \bx[b]) = \norm{\by[a] - \by[b]}_1.$%
    \footnote{%
        The notation $\bb{a}\i[i]$ is alternative notation for either a named variable or the $i^\text{th}$ element of the ordered set $\set{A}$. %
    }
    Similarly, we define the \textit{perceived distance} as the difference between the two observations in pixel space as measured by the model's reconstruction loss function (usually MSE): $\VDistance(\bx[a], \bx[b]) = \LossRec(\bx[a], \bx[b]).$%
    \footnote{We use the term increased \textit{perceived overlap} as a synonym for lower \textit{perceived distance} between neighbouring states.}

\begin{figure}
            \centering
            \begin{subfigure}{0.59\textwidth}
                \centering
                \includegraphics[width=0.72\linewidth]{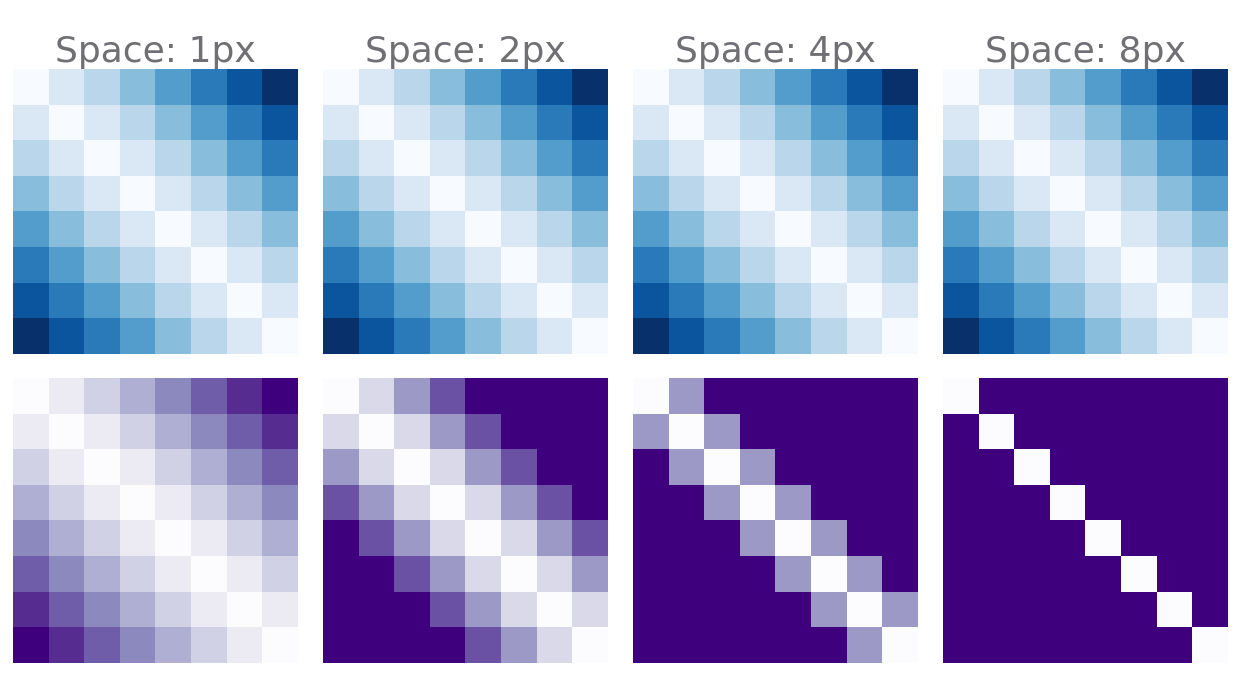}
                \caption{Top Row: Manhattan ground-truth distance matrices over factor traversals. Bottom Row: Pixel-wise perceived distance matrices between observations over factor traversals. Left to Right: Step size increases from 1px to 8px. Smaller values result in more overlap in the data space, which leads to higher a probability of being able to find an ordering along a factor traversal.}
                \label{fig:data-dists_xysquares-incr-overlap}
            \end{subfigure}
            \hfill
            \begin{subfigure}{0.385\textwidth}
                \centering
                \includegraphics[width=1.0\linewidth]{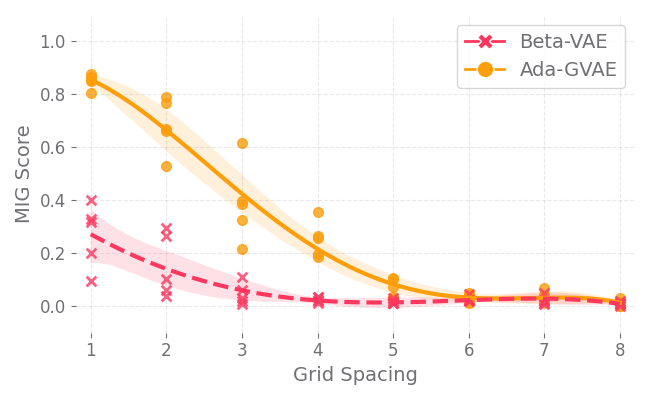} 
                \caption{
                   Regression plot of increasing domain spacing versus MIG score (higher is better). As the $\beta$-VAE (dashed line) and Ada-GVAE (solid line) are trained with decreasing (left to right) levels of overlap, their disentanglement performance worsens. %
                }
                \label{fig:exp-incr-overlap}
            \end{subfigure}
            \caption{Varying spacing of valid locations in gridworld domain affects overlap as perceived by an VAE.}
        \end{figure}

    To investigate the above question, we modify the environment so that the agent moves in smaller increments. 
    For each step size $s$, we generate all possible image states and store them in a buffer.\footnote{We modify the size of the environment to ensure the size of the buffer remains constant regardless of the step size.} 
    As $s$ decreases, the probability that the agent position overlaps in any two randomly sampled images increases.
    We next visualise ground-truth and perceived distances for each step size setting, with the results given by \Cref{fig:data-dists_xysquares-incr-overlap}.
    The results show a clear relationship between the two distance measures---at very small step sizes (with high probability of overlap) there is almost perfect correspondence.
    
    We further probe this relationship by using the data buffers as an unsupervised dataset. 
    We make the problem harder by converting it to a multiagent environment which increases the number of states---three objects are now each described by their $xy$-positions.
    We train a $\beta$-VAE \citep{higgins2017beta} and the state-of-the-art weakly-supervised Ada-GVAE \citep{locatello2020weakly}. 
    The $\beta$-VAE scales the VAE regularisation term with a coefficient $\beta > 0$, while the Ada-GVAE encourages axis alignment and shared latent variables between pairs of observations. This is achieved by averaging together latent distributions between observation pairs that are estimated to remain unchanged when the KL divergence is below some threshold.
    To evaluate if the representations are disentangled, we use the Mutual Information Gap (MIG) \citep{chen2018isolating} 
    The results in \Cref{fig:exp-incr-overlap} indicate that as the spacing decreases and overlap is introduced, the disentanglement performance improves as perceived pixel-wise distances better correspond to latent and ground-truth distances. %

\section{Triplet Loss} \label{section:triplet}
    
    VAEs, therefore, appear to learn latent distances over the data that correspond to their reconstruction loss. However, if these distances are not useful,
    we may be able to instead guide the learning process 
    by introducing \textit{metric} or \textit{similarity} learning.
    A common approach is triplet loss, which makes use of three observations $\X_{\mathrm{triple}} = (\bx[a], \bx[p], \bx[n])$. These are sampled using supervision such that the \textit{anchor} $\bx[a]$ is considered closer to \textit{positive} $\bx[p]$ than it is to the \textit{negative} $\bx[n]$. The intuition is that corresponding representations $\bz[a], \bz[p], \bz[n]$ output by a network should satisfy the original anchor-positive and anchor-negative distance constraints: $d(\bz[a], \bz[p]) < d(\bz[a], \bz[n])$.
    We construct the $\beta$-TVAE as three parallel versions of the same $\beta$-VAE augmented by the soft-margin formulation of triplet loss, where $\alpha > 0$ is the scaling factor: 
    \begin{equation}
        \LossTVAE =
        \alpha  \underbrace{\ln \parens{
            1 + \exp \parens { \Dist(\bz\i[a], \bz\i[p]) - \Dist(\bz\i[a], \bz\i[n]) }}}_{\text{Triplet loss}}
        + \frac{1}{3} \sum_{\bx \in \X_{\mathrm{triple}}} \LossBVAE(\bx)
        \label{eq:triplet_vae_loss}
    \end{equation}

    \subsection{Adaptive Triplet Loss} \label{sec:axis-aligned-triplet}
    
        Standard triplet loss provides no direct pressure to learn factored representations;
        for example the $x$ and $y$ factors may be encoded with some arbitrary rotation in the latent space.
        To fix this, we adapt the Ada-GVAE\citep{locatello2020weakly} to construct Ada-Triplet and the Ada-TVAE. These adaptive methods encourage partially shared representations such that differences between observations are encoded in subsets of latent units.
            Shared latent units are estimated as those whose distances $\delta_i = |z_i\i[a] - z_i\i[n]|$ are less than half way between the maximum and minimum: $\delta_i < \frac{1}{2}\parens{\min_{i}\delta_i + \max_{i}\delta_i}$. To encourage factored latents, we elementwise multiply $\elemprod$ arguments of the anchor-negative triplet term by the weight vector $\boldsymbol{w} = (\omega_1, ..., \omega_\sZ)$. Shared latents are weighted less $\omega_i \in (0, 1)$ and non-shared units $\omega_i = 1$ remain unchanged:
        \begin{equation}
            \LossAdaTrip = \ln \parens{
                1 + \exp \parens { \Dist(\bz[a], \bz[p]) - \Dist(\boldsymbol{\omega} \elemprod \bz[a], \boldsymbol{\omega} \elemprod \bz[n]) }
            } \label{eq:ada_triplet_loss}
        \end{equation}

    \subsection{Using the Replay Buffer for Metric Learning and Feature Extraction} \label{sec:triplet_random_walk_sampling}
        
        Typically, triplet loss is used in the supervised learning case, where the anchor, positive and negative samples are known. 
        However, no such signal exists in the RL setting. 
        To overcome this, we leverage the sequential nature of the problem and assume that \textit{states closer together in time are also on average closer in terms of their ground-truth factors.}

\begin{figure}[h!]
            \centering
            \begin{subfigure}[b]{.6\textwidth}
                \centering
                \includegraphics[width=1.0\linewidth]{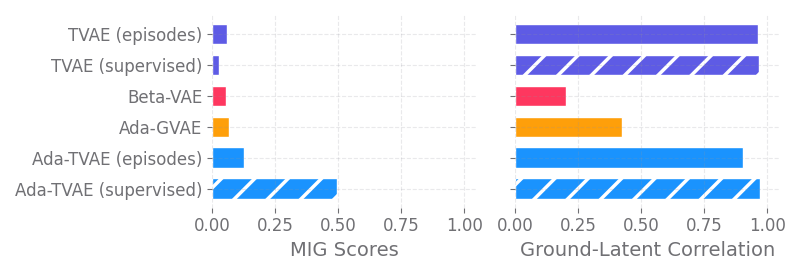}
                \caption{} \label{fig:exp-random-walk-scores}
            \end{subfigure}
            \hfill
            \begin{subfigure}[b]{.35\textwidth}
                \centering
                \includegraphics[width=1.0\linewidth]{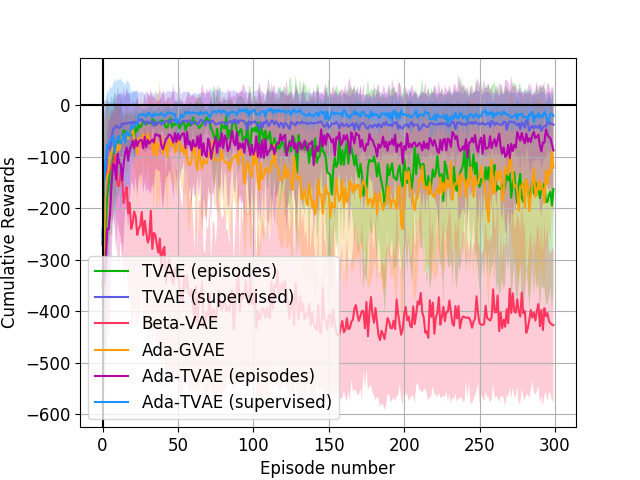} 
            \caption{}  \label{fig:q_learning}
            \end{subfigure}
            \caption{
                (a) Different VAE frameworks trained to extract features. %
                The plot shows the average rank correlation between the ground-truth distances and corresponding distances between these learnt features. 
                Striped bars are models trained using ground-truth distances, while solid bars use our approach of sampling sequentially from the replay buffer.
                (b) Downstream RL training rewards using previously learnt features. Mean and standard deviation are given over 30 runs.}
        \end{figure}

        Given a replay buffer $\mathcal{B} = \{s_0, a_0, r_0,  s_1,a_1, r_1, \ldots  \}$, we use this assumption to sample some anchor state $s_a$, a positive $s_p$  and a negative $s_n$ such that $n > p > a$.
        To test the effectiveness of this assumption, we execute a uniformly random policy to populate a replay buffer. 
        We train a $\beta$-VAE and Ada-GVAE as baselines, which we compare against the $\beta$-TVAE and Ada-TVAE from the previous sections.
        As a sanity check, we also train a fully supervised $\beta$-TVAE and Ada-TVAE on the underlying ground-truth factors of the environment (the agent's $xy$-position) so that the anchor-positive $\ell_1$ distance is always less than the anchor-negative.
        The results in \Cref{fig:exp-random-walk-scores} demonstrate that triplet-based models significantly outperform the baselines by learning better features.
    
        Finally, we test the effect of using the encoders of these learnt models as feature extractors in an RL setting. 
        We freeze the weights of the previously trained encoders, and apply deep Q-learning \citep{mnih2015human} over these features to solve the task of reaching the top left corner from the bottom right (rewards of $-1$ on all timesteps).
        Results in \Cref{fig:q_learning} demonstrate that features from the adaptive triplet loss model are the most useful for the RL task; however, this is well approximated by sampling triplets using only the ordering of states from the replay buffer. Both triplet methods outperform the baselines; however, adaptive triplet encourages factored representations which are better features for the downstream RL task.
        
\section{Conclusion}
    
    We identified a shortcoming of VAE-based approaches in domains where there is little or no variation in perceived distances or overlap. 
    This may be inherent in the environment, or arise inadvertently due to design decisions. 
    For example, if frame skipping \citep{mnih2015human} or high-level skills are used, there may be no perceived overlap between successive states in the buffer.
    We overcome this problem by incorporating the inherent temporal information present in an episode through the simple use of metric learning. Furthermore, we improve the performance by enabling representation disentanglement by designing an adaptively weighted version of triplet loss.
    Leveraging the sequential nature of the replay buffer to perform metric learning may open up interesting new avenues for future feature learning approaches in RL.

{
\footnotesize
\bibliography{main}
}

\end{document}